\documentclass[sigconf]{acmart} 

\acmConference[RCE Exam]{Research Competency Evaluation Exam}{April 17, 2026}{Houston, Texas, USA}

\usepackage{preamble}
\title{The Collection of a Real-World, Large-Scale Dataset for \\ Automated Skill Evaluation in Central Line Dressing Change}

\begin{abstract}
    Central line-associated bloodstream infections (CLABSIs) remain a major source of preventable morbidity and mortality in healthcare settings. 
Proper execution of the central line dressing change (CLDC) procedure is essential for infection prevention, yet current approaches to competency assessment rely primarily on manual observation by clinical educators, making evaluation time-intensive, difficult to scale, and limited in consistency and granularity. 
These challenges highlight the need for AI-assisted skill evaluation methods that can support more objective, reproducible, and scalable assessment of clinical performance. 
However, progress in this area has been constrained by the lack of large-scale, clinically grounded datasets for developing and evaluating such methods.

In this work, we present the collection process of a large-scale video dataset for CLDC skill evaluation, comprising 365 recordings of registered nurses collected during the 2026 annual competency exam at Houston Methodist Hospital. 
Each video is paired with subtask-level temporal annotations and clinical performance assessments spanning 17 subtasks of the procedure. 
We describe the data collection and curation pipeline, including pilot testing of recording configurations, annotation protocol design, and post-collection processing. We further establish the feasibility of video-based skill assessment through benchmark experiments with spatiotemporal deep learning models. 
This dataset provides a foundation for future research on automated clinical skill assessment, feedback generation, and AI-assisted training in nursing education.

\end{abstract}

\begin{document}
%%%%%%%%%%%%%%%%%%%%%%%%%%%%%%%%%%%%%%%%%%%%%%%%%%%%%%%%%%%%%%%%%%%%%%%%
%%% The following commands remove the headers in your paper. For final 
%%% papers, these will be inserted during the pagination process.
\pagestyle{fancy}
% \fancyhead{}
%%% The next command prints the information defined in the preamble.

%Automating_Skill_Evaluation_Teaser
\begin{teaserfigure}
  \centering
  \includegraphics[width=0.8\textwidth,keepaspectratio]{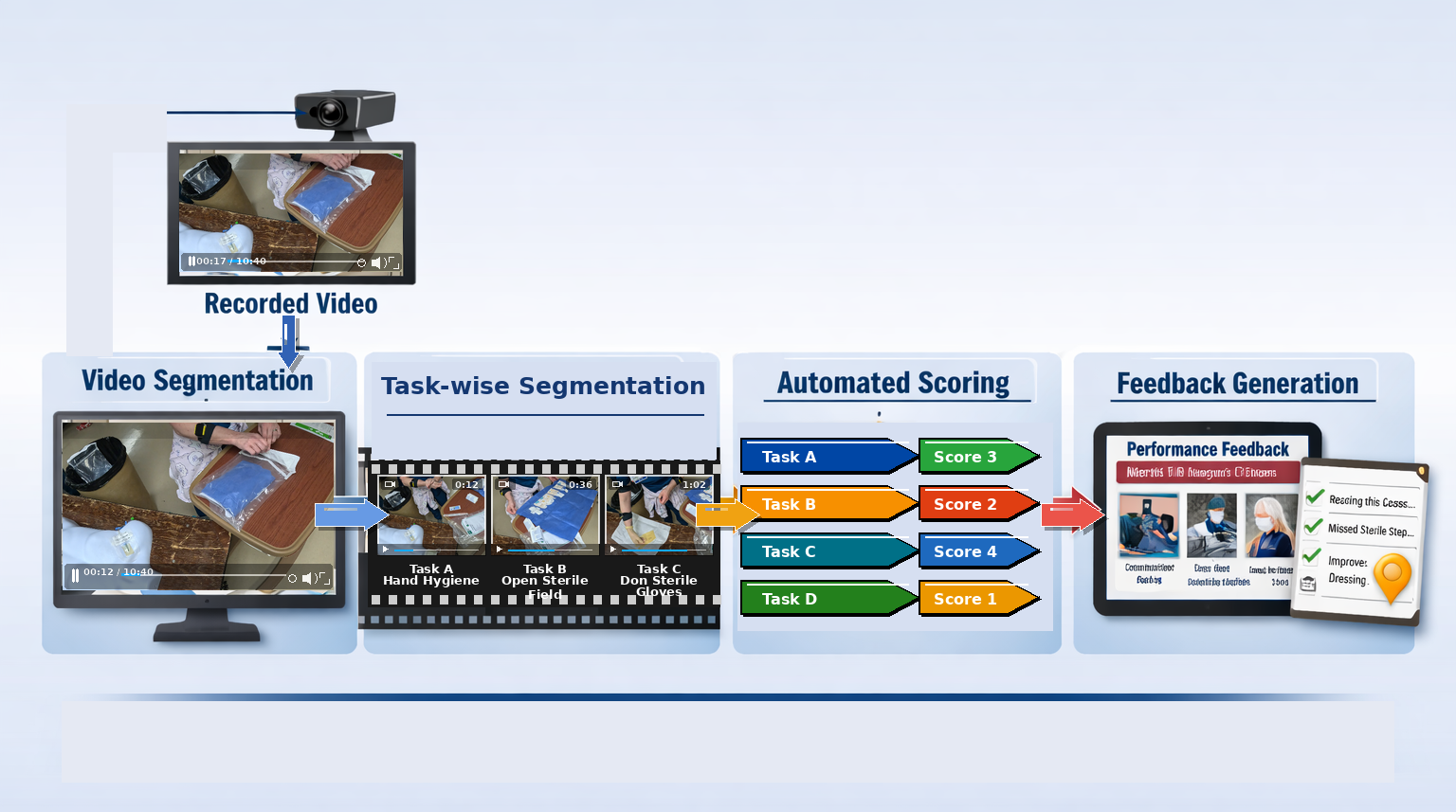}
  \caption{Overview of the proposed AI-assisted pipeline for central line dressing change evaluation. Recorded evaluation videos are segmented into task-specific clips, scored automatically at the subtask level, and used to generate performance feedback.}
  \Description{describe the image.}
  \label{fig:teaser}
\end{teaserfigure}
%%%%%%%%%%%%%%%%%%%%%%%%%%%%%%%%%%%%%%%%%%%%%%%%%%%%%%%%%%%%%%%%%%%%%%%%
\maketitle 
%%%%%%%%%%%%%%%%%%%%%%%%%%%%%%%%%%%%%%%%%%%%%%%%%%%%%%%%%%%%%%%%%%%%%%%%

\section{Introduction}
\label{sec:introduction}

\begin{figure}[H]
   \includegraphics[width=6cm,height=3cm]{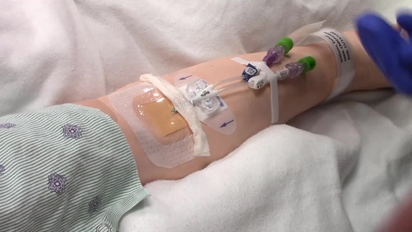}
   \caption{Example of a central line dressing, illustrating the dressing, central lines, and securement device.}
   \label{fig:CLDC}
\end{figure}

Central line-associated bloodstream infections (CLABSIs) represent one of the most severe and costly forms of healthcare-associated infections, with reported mortality rates ranging from 12\% to 25\% and an estimated 28,000 deaths occurring annually in U.S. intensive care units alone \cite{AHRQ2024CLABSI, Haddadin2023Central, Star2024Reducing}. 
Evidence demonstrates that patients with CLABSI face a mortality odds ratio of 3.19 compared to uninfected patients, alongside an average of 16 additional days of hospitalization and costs of approximately \$45,000--\$46,000 per case \cite{Elangovan2024Impact, Star2024Reducing, Haddadin2023Central}. 
Individual hospital-level studies further confirm that patients who develop a CLABSI are 36.6\% more likely to die and 37.0\% more likely to be readmitted than those with central venous catheters who remain infection-free \cite{Chovanec2021Association}. 
These findings collectively demonstrate that CLABSIs impose a substantial and measurable burden on patient outcomes and healthcare systems. 
A significant contributing factor to CLABSI incidence is inconsistency in how nurses perform the CLDC procedure, which is a complex, multi-step clinical task that demands strict adherence to sterile technique and procedural accuracy.
The CLDC procedure involves removal of the old dressing, maintenance of a sterile field, cleansing of the insertion site, and application of new dressing materials to protect the catheter site and reduce infection risk. 
Because the procedure requires careful sequencing of subtasks and sustained compliance with infection prevention protocols, even small deviations in technique may compromise sterility and increase patient risk. 
An example of a central line dressing setup is shown in Figure~\ref{fig:CLDC}.

Current methods for evaluating nursing competency in CLDC rely predominantly on direct observation by clinical educators, a process that is time-intensive, subjective, and difficult to scale across large healthcare institutions \cite{gorski2021}. 
These limitations motivate the need for AI-assisted skill evaluation methods, illustrated in Figure~\ref{fig:teaser}, that can support more objective, reproducible, and scalable assessment of clinical performance. 
However, progress in this area has been constrained by the lack of large-scale, clinically grounded datasets for nursing skill evaluation. 
Existing efforts in this space have largely focused on simple rule-based skill tracking and monitoring concepts, rather than data-driven video-based evaluation of real clinical performance \cite{qian2025astrid}. 
This limitation reflects a broader data bottleneck: \textbf{without sufficiently large and well-annotated datasets, it remains difficult to develop and evaluate robust AI methods for fine-grained assessment of nursing procedures such as CLDC.}

To address this gap, this paper introduces a large-scale video dataset for CLDC skill evaluation, collected during authentic competency evaluations at Houston Methodist Hospital. 
The dataset comprises 365 recordings of registered nurses and is paired with structured temporal annotations and clinical performance assessments across key subtasks of the procedure. 
By establishing a real-world dataset for CLDC skill evaluation, this work provides foundational data infrastructure for future research on automated procedural understanding, AI-assisted skill assessment, and feedback generation in nursing education.
\section{Related Work}
\label{sec:related-work}

Our review of prior literature spans three areas most relevant to this work: procedural video datasets, automated video-based skill assessment, and simulation-based clinical nursing evaluation. 
Across these areas, prior work has established the value of annotated procedural data and machine learning methods for skill analysis, but no existing study has directly addressed automated skill evaluation for central line dressing change using real nursing video collected in authentic competency assessment settings.

\subsection{Healthcare Procedural Video Datasets}

The development of large-scale, annotated video datasets has become an important foundation for AI-driven healthcare training and procedural analysis, although the field remains constrained by limited scale, narrow procedural focus, and inconsistent labeling practices \cite{huNurViD2023, slamDataset2025, surgicalToolSurvey2022}. 
NurViD \cite{huNurViD2023} addressed part of this gap by introducing more than 1,500 publicly available nursing videos collected from open-source YouTube instructional materials, spanning 51 procedures and 177 action steps for procedure recognition and action detection. 
In the surgical domain, the SLAM dataset \cite{slamDataset2025} provided 4,097 annotated laparoscopic video clips and demonstrated validation with Video Vision Transformers achieving 85.90\% accuracy, while SurgVU \cite{surgVU2025} enabled weakly supervised methods for activity recognition and skill assessment. 
A comprehensive survey of surgical tool datasets \cite{surgicalToolSurvey2022} further cataloged available data resources and associated machine learning techniques across robotic surgery, instrument tracking, and pose recognition. 
More broadly, a recent review \cite{DLSurgicalWorkflow2024} systematically analyzed deep learning methods for surgical workflow recognition from 2018 to 2024, underscoring the importance of richly annotated video datasets for deploying computer vision in healthcare training applications. 
However, these datasets remain focused primarily on surgical or instructional settings rather than real-world nursing competency evaluation.

\subsection{Video-Based Automated Skill Assessment}

Automated and objective assessment of medical and surgical skill has advanced rapidly through the application of computer vision and deep learning to procedural video \cite{lavanchyAutomation2021, hashemiVideoRAS2025, jamaNotss2024, powerLSPD2025}. 
Lavanchy et al. \cite{lavanchyAutomation2021} presented a three-stage pipeline consisting of instrument detection, motion feature extraction, and regression-based skill prediction, achieving 87\% accuracy in distinguishing good from poor laparoscopic performance. 
Hashemi et al. \cite{hashemiVideoRAS2025} extended this direction with a Convolutional Neural Network(CNN)-Long Short Term Memory (LSTM) architecture for simultaneous action recognition and skill classification across 130 robot-assisted surgical videos, while a JAMA Surgery study \cite{jamaNotss2024} showed that OpenPose-derived team motion features from cardiac surgery videos correlate with nontechnical skills assessed using the Non-Technical Skills for Surgeons (NOTSS) tool. 
Sensor-free approaches have also emerged. 
For example, the KCF-ResNet framework \cite{kcfResNet2023} extracted motion signals directly from video without physical instrument sensors, and a dual AI framework combining ResNet-50 segmentation with YOLOv2 tracking \cite{sugiyamaMicro2025} demonstrated strong multimodal skill discrimination in microsurgery.

A recent systematic review \cite{frontiersSurgReview2025} concluded that AI-based video analysis can provide more objective skill evaluation than manual assessment while reducing bias and cost. 
Recent work has further extended automated assessment to open surgery through egocentric video analysis using Transformer and Temporal Convolutional Network (TCN) architectures with OSATS-scale (Objective Structured Assessment of Technical Skills) annotations \cite{egocentricSkill2025}, and to simulated settings through the LSPD dataset \cite{powerLSPD2025}. 
Collectively, these studies demonstrate the feasibility of video-based automated skill assessment, but they focus largely on surgical tasks performed in controlled operative or simulation environments.

\subsection{Clinical Nursing Skill Evaluation}

Clinical nurse evaluation has evolved substantially, with simulation-based approaches and structured assessment instruments increasingly supplementing traditional preceptor-led observation in order to address growing student populations, ethical constraints, and the persistent theory--practice gap \cite{bridgingGapBMC2025, sblSystematic2024, chabreraOSCE2023, osceKorea2023}. 
Systematic reviews \cite{sblSystematic2024, bridgingGapBMC2025} have shown that simulation-based learning across varying fidelity levels, from low-fidelity mannequins to high-fidelity simulators, can improve knowledge and skill acquisition, although evidence regarding long-term retention remains limited. 
The Objective Structured Clinical Examination (OSCE) has become
a prominent standardized evaluation strategy. Validation studies
have reported content validity indices (CVI) of .82--.95,
indicating that 82--95\% of expert reviewers rated the examination
content as relevant and representative of the intended clinical
skills. Additionally, reliability coefficients of $\alpha = .84$
(Cronbach's alpha) suggest strong internal consistency across
examination items, meaning the OSCE stations consistently measure
the same underlying clinical competency \cite{chabreraOSCE2023}. 

Korean studies further showed that OSCE implementation after mannequin-based practice significantly improved knowledge retention across core nursing skills \cite{osceKorea2023}. 
At the same time, research on simulation design elements \cite{simDesignCSN2024} has shown that omission of key components such as prebriefing, fidelity alignment, and structured debriefing can impair learning outcomes, supporting alignment with  the International Nursing Association for Clinical Simulation and Learning (INACSL) best-practice frameworks. 
Qualitative work has also found that students experience challenges with mannequin realism and that inadequate technological infrastructure can exacerbate evaluation-related stress \cite{osceStressPLOS2025}. 
Although immersive VR has been explored as an alternative modality providing multi-sensory clinical immersion, most evidence remains focused on learner satisfaction rather than objective skill measurement \cite{vrScopingCSN2024}. 
In addition, studies using scaffolded OSCEs have found that learner satisfaction and self-confidence correlate only weakly with actual performance grades \cite{sibleyOSCE2025}, underscoring the continued need for objective, competency-based evaluation methods in nursing education.

Taken together, the literature highlights a clear gap. 
Existing surgical datasets and automated skill assessment models demonstrate the promise of video-based evaluation, while existing nursing education studies continue to rely primarily on instructor-led, checklist-based assessment. 
No prior work, to our knowledge, has introduced a large-scale real-world video dataset specifically designed for automated skill evaluation of the central line dressing change procedure. 
Our work addresses this gap by introducing a large-scale CLDC dataset collected during authentic nursing competency evaluations and paired with structured temporal annotations and clinical performance labels, thereby providing a foundation for future development of automated assessment methods in nursing education.
\section{Materials and Methods}
\label{sec:methods}

The study was designed as a multi-phase research effort encompassing benchmark experimentation, clinical task definition, data collection protocol development, pilot recording, real-world clinical data collection, and annotation workflow design.

\subsection{Benchmark Validation on JIGSAWS}

Before collecting domain-specific nursing data, we replicated and extended existing video-based skill assessment methods on the JIGSAWS Surgical Activity Dataset \cite{funke2019video} to validate our computational pipeline. 
JIGSAWS contains recordings of eight surgeons with varying levels of experience, from novice to expert, performing three standardized tasks (Suturing, Knot-Tying, and Needle-Passing) across 103 usable trials. 
We evaluated two model architectures using RGB video only, with 5-fold cross-validation applied across all experiments.

\begin{table}[h]
\centering
\caption{Benchmark replication results on JIGSAWS.}
\label{tab:jigsaws}
\small
\begin{tabular}{p{2.8cm}ccc}
\hline
\textbf{Paper / Model} & \textbf{Data} & \textbf{Model} & \textbf{Train Accuracy} \\
\hline
Funke et al. (2019) & RGB & I3D + TSN & 95.1--100\% \\
Replicated Model 1 & RGB & I3D + TSN & 95.45\% \\
Replicated Model 2 & RGB & C3D + TSN & 90.12\% \\
\hline
\end{tabular}
\end{table}

Our first replicated model paired an Inflated 3D ConvNet (I3D) with a Temporal Segment Network (TSN), following the approach described by Funke et al. \cite{funke2019video}, who reported training accuracies ranging from 95.1\% to 100\% across the three tasks. 
Our I3D + TSN replication achieved an overall training accuracy of 95.45\%, closely matching the original results and confirming that our training pipeline, data preprocessing, and evaluation protocol were correctly implemented. 
We also trained a second model in which the I3D backbone was replaced with a Convolutional 3D Network (C3D) network integrated with TSN, which yielded a lower overall accuracy of 90.12\%. 
This performance gap suggests that the richer spatiotemporal representations learned by I3D provide a meaningful advantage over the shallower temporal features captured by C3D for fine-grained skill classification.

% \begin{figure}[H]
%     \centering
%     \includegraphics[width=5cm]{images/TaskwiseSkillAcconJIGSAWS.png}
%     \caption{Task-wise skill accuracy on JIGSAWS.}
%     \label{fig:TaskwiseAccJIgsawa}
% \end{figure}

% Following the original study, we also found that Suturing consistently achieved the highest accuracy across both architectures (82\% with I3D + TSN and 79\% with C3D + TSN), likely because suturing involves distinctive and repetitive hand motions that are easier for the model to associate with different skill levels. Knot-Tying and Needle-Passing yielded comparable but slightly lower accuracies (78--80\% across both models), reflecting the greater variability in how surgeons of different skill levels execute these tasks. The overall task-wise trend remained consistent across both architectures, with the I3D-based model maintaining a performance advantage. These benchmark results provided an initial validation of our video-based skill assessment pipeline before applying related modeling ideas to the CLDC domain.

\subsection{Defining CLDC Subtasks}

% \begin{figure}[H]
%     \centering
%     \includegraphics[width=8cm]{images/BreakCLDCintoUniqueSubtasks.png}
%     \caption{Identifying unique subtasks in the CLDC procedure.}
%     \label{fig:Subtask}
% \end{figure}

To enable automated skill evaluation, the CLDC procedure must first be decomposed into discrete subtasks that support temporal annotation, step-level assessment, and fine-grained modeling of clinical actions from video.
In consultation with clinical experts at Houston Methodist, the CLDC procedure was decomposed into a finalized set of 19 discrete subtask activities (A01 through A19), following the hospital-provided checklist for central line dressing change as shown in Table~\ref{tab:cldc_activities}.
These activities span the full procedure. The activity taxonomy was iteratively refined based on observations during nursing classes at Houston Methodist and feedback from clinical educators. 
The StatLock-related activities (A06, A14) and activity A01\_RemoveContents were retained in the taxonomy for completeness, although they were noted as not performed during the annual competency exam.

\begin{table}[ht]
\centering
\caption{Finalized subtasks for the CLDC dataset }

\label{tab:cldc_activities}
\begin{tabular}{ll}
\hline
\textbf{Activity ID} & \textbf{Activity Description} \\
\hline
A01\_RemoveContents        & Remove contents of the pocket \\
A02\_HeadCoverMask         & Head cover and masks \\
A03\_PerformHandHygiene    & Perform hand hygiene \\
A04\_NonSterileGloves   & Non-sterile gloves \\
A05\_DressingRemoval         &  Detachol to remove dressing \\
A06\_RemoveStatlock        & Remove statlock \\
A07\_DisposeEverything     & Dispose everything \\
A08\_OpenSterileField      & Open sterile field \\
A09\_DonSterileGloves      & Don sterile gloves \\
A10\_AlcoholSwabstick      & Apply alcohol swabstick \\
A11\_CHGSponge             & Apply CHG sponge \\
A12\_ApplyCavilonBarrier   & Apply cavilon barrier \\
A13\_MastisolApplication   & Apply mastisol \\
A14\_ApplyStatlock         & Apply statlock \\
A15\_ApplyCHGDressing      & Apply CHG dressing \\
A16\_ApplyBifurcatedPants  & Apply bifurcated pants \\
A17\_ApplyBifurcatedBelts  & Apply bifurcated belts \\
A18\_PlaceLabelOnDressing  & Place label on dressing \\
A19\_PlaceClearTape        & Place clear tape \\
\hline
\end{tabular}
\end{table}

%donning head cover and mask (A02), donning non-sterile gloves (A03), removing the old dressing (A04), and disposing of soiled materials (A05), through opening the kit and placing items on the sterile surface (A06), cleansing the site with alcohol swabsticks (A08), applying Cavilon barrier (A09), applying Mastisol adhesive (A10), applying dressing layers (A11), . \pq{List all subtasks in a table.} %

This task decomposition served as the basis for subsequent annotation, evaluation, and dataset organization. The central research question was therefore formulated as follows: How can we design and implement a video-based automated system for evaluating step-level nursing skill in the central line dressing change procedure by combining real-world clinical video data with state-of-the-art action recognition and skill classification methods?

\subsection{Establishing Clinical Collaboration}

A critical enabler of this research was the establishment of a close collaborative relationship with Houston Methodist Hospital, which served as both the clinical domain partner and the primary site for data collection. 
The collaboration was initiated through discussions led by Prof. Qian, who engaged clinical stakeholders at Houston Methodist to explore the feasibility and mutual benefit of the project.

Houston Methodist expressed strong interest in the research from the outset. 
The hospital had previously worked with an external company to record nursing procedures for manual review by instructors. 
However, that initiative ultimately did not move forward, and no existing video recordings were available from that collaboration. 
Importantly, the experience had revealed a key limitation: instructors found that manually reviewing video recordings was more burdensome than in-person observation because the assessment process was not automated. 
This prior experience made Houston Methodist highly receptive to the idea of automated skill assessment, and the hospital recognized the potential for AI-driven evaluation to address the scalability and efficiency challenges that had limited earlier efforts.

Houston Methodist provided institutional support for the research team to collect data on-site. 
This included permission for the research team to observe CLDC training classes, which are offered every other Wednesday, and to attend the annual January nursing competency evaluations, a large-scale assessment event in which approximately 2,000 nurses are evaluated on the CLDC procedure across multiple stations over a multi-week period. 
We targeted a minimum of 300 video recordings from these sessions in order to build a dataset large enough to support the training and evaluation of video-based skill assessment models. 
The hospital also facilitated introductions between the research team and the clinical educators who oversee the competency evaluations. 
In preparation for data collection at Houston Methodist, several logistical and ethical considerations were addressed jointly with hospital leadership. 
The research team prepared an Institutional Review Board (IRB) application to Houston Methodist and agreed not to include audio or facial imagery in the dataset. 
This close clinical partnership was essential for ensuring that the research remained grounded in authentic clinical practice and that the resulting dataset reflected the real-world complexity of nursing skill evaluation.

\subsection{Pilot Recordings and Camera Configuration}

A series of pilot recording sessions was conducted in the university lab to test and refine the data recording setup before deployment in the clinical environment. 
During these sessions, we replicated the central line dressing change setup and practiced performing the CLDC procedure to generate pilot recordings from team members. 
This process allowed us to evaluate video quality across different setups and identify the recording configuration and equipment best suited for capturing the fine-grained hand movements and toolkit interactions that distinguish one subtask from another.

\vspace{-10pt}
\begin{table}[h]
\centering
\caption{Cameras evaluated during pilot recording sessions.}
\label{tab:cameras}
\small
\begin{tabular}{p{2.2cm}ccc}
\hline
\textbf{Camera} & \textbf{Resolution} & \textbf{FOV} & \textbf{FPS} \\
\hline
RealSense D435i & 1280$\times$720 & 87$^\circ \times 58^\circ$ & 30 \\
GoPro HERO13 & 4K & 156$^\circ$ & 60 \\
4K Camcorder & 4K & Ultra-wide & 30 \\
\hline
\end{tabular}
\end{table}

%\paragraph{Choosing Cameras} 
We tested three cameras side by side, as summarized in Table~\ref{tab:cameras}. The Intel RealSense D435i offered RGB-D capture, which we initially considered useful for depth-based hand tracking, but its 1280$\times$720 resolution produced blurry renderings of the text printed on kit labels and swab packaging, causing important visual cues to be lost. 
The 4K Ultra HD camcorder delivered sharp footage with an ultra-wide field of view, but its bulkier form factor made it difficult to mount unobtrusively in a clinical room. 
The GoPro HERO13 Black struck the best balance: it recorded at 4K resolution with up to 60 FPS, offered a 156$^\circ$ field of view at its widest setting, and was compact enough to attach to a standard tripod without interfering with the participant's workspace.
\captionsetup{justification=centering}
\begin{figure}[H]
    \centering
    \includegraphics[width=5cm]{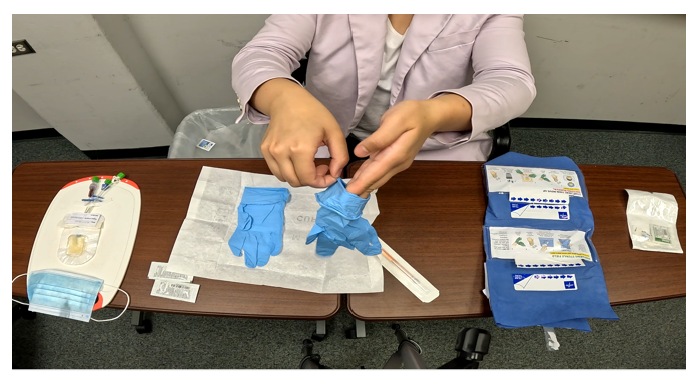}
    \caption{Pilot recordings in the lab using angled top-down center view.}
    \label{fig:PilotRecordings}
\end{figure}

\begin{figure}[H]
    \centering
    \includegraphics[width=8cm]{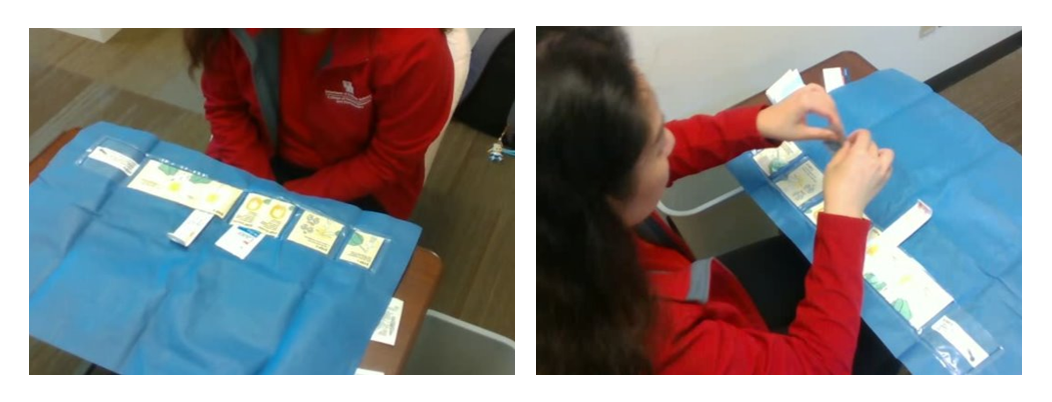}
    \caption{Other camera mounting positions tested during pilot recordings: (left) frontal-left top-down corner; (right) corner overhead diagonal.}
    \label{fig:CameraAngles}
\end{figure}

%\paragraph{Testing Recording Angle} 
We also experimented with three camera mounting angles: (1) an angled top-down center view, positioned directly above and in front of the participant; (2) a corner overhead diagonal view, mounted at the top-right corner of the table and oriented downward at an angle; and (3) a frontal-left top-down view, placed slightly to the left of the participant above the front corner of the work surface. Figure~\ref{fig:PilotRecordings} demonstrates (1) and Figure~\ref{fig:CameraAngles} demonstrates (2) and (3); the final recording uses (1).
For each angle, we reviewed the resulting footage to assess visibility of the sterile field, the extent to which the operator's hands occluded critical actions, and the consistency of framing across different people performing the procedure.

\captionsetup{justification=centering}
% We also noted the start and end timestamps of each subtask during the pilot recordings to test whether our annotation workflow was practical at the pace of a real procedure. 

Based on these pilot experiments, the selected configuration mounted the camera on a tripod at a height of approximately 5'4'' to 5'5'' above the floor. 
The camera was angled 50--60$^\circ$ downward toward the procedure area, providing an angled top-down view that captured the full sterile field, the participant's hands and upper torso, and the dressing toolkit. This configuration balanced the need for comprehensive coverage of the procedural workspace with minimization of facial exposure for privacy purposes.

% \vspace{-5cm}

\subsection{Preparation for Recording at Houston Methodist}

In preparation for data collection at Houston Methodist, we undertook several organizational and logistical steps to ensure that the recording sessions would run smoothly and yield consistent, high-quality data. 
We first prepared the materials required for the recording protocol. 
A demographic survey was developed to capture information about each participant's background, including age, sex, years of nursing experience, nursing specialty, work shift, and self-reported confidence with the CLDC procedure.

\begin{figure}[H]
    \centering
    \includegraphics[width=8cm, height=8cm]{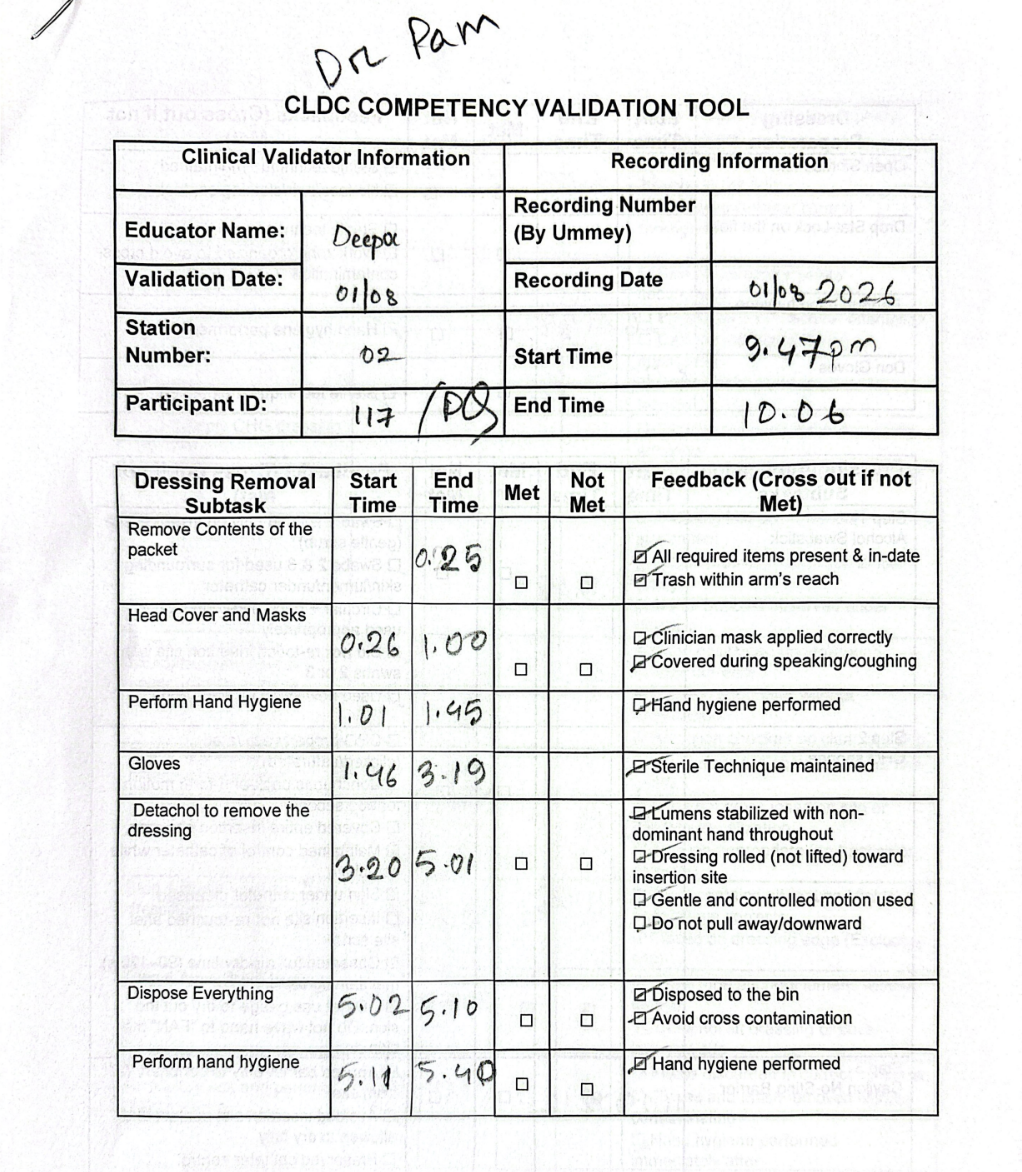}
    \caption{Printed data annotation sheet used during live recording sessions at Houston Methodist.}
    \label{fig:TimeAnnotationSheet}  \setlength{\intextsep}{0pt}
\end{figure}

We also designed a printed data annotation sheet (Figure~\ref{fig:TimeAnnotationSheet}) containing the list of feedback items required for CLDC evaluation during each session. 
Printed data annotation forms were selected over digital alternatives after pilot testing showed that paper forms were faster and more reliable for recording timestamps in a fast-paced environment. 
These forms allowed a member of the research team to record subtask-level start and end timestamps while a clinical nursing educator simultaneously assessed the participant's performance against predefined quality indicators. 
In addition, we designed a separate online evaluation form for team members to record performance assessments and feedback independently of the printed annotation sheets used during recording. Participant ID cards were prepared in advance and displayed to the camera at the start of each recording in order to associate the participant ID with the corresponding recording and annotation forms.

% \begin{figure}[H]
%     \centering
%     \includegraphics[width=8cm,height=5cm]{images/OnlineEvaluationForm.png}
%     \caption{Online CLDC evaluation form used for structured performance assessment.}
%     \label{fig:CLDCEvaluationForm}
% \end{figure}
\captionsetup{justification=centering}
The Houston Methodist team also coordinated with hospital staff to determine permissible camera placement locations within the competency evaluation rooms, secure access to charging ports for camera batteries, and confirm scheduling details for each day of the three-week data collection period. Hospital leadership and clinical educators played an active and supportive role throughout the preparation process.
\captionsetup{justification=centering}
\begin{figure}[H]
    \centering
    \includegraphics[width=7cm]{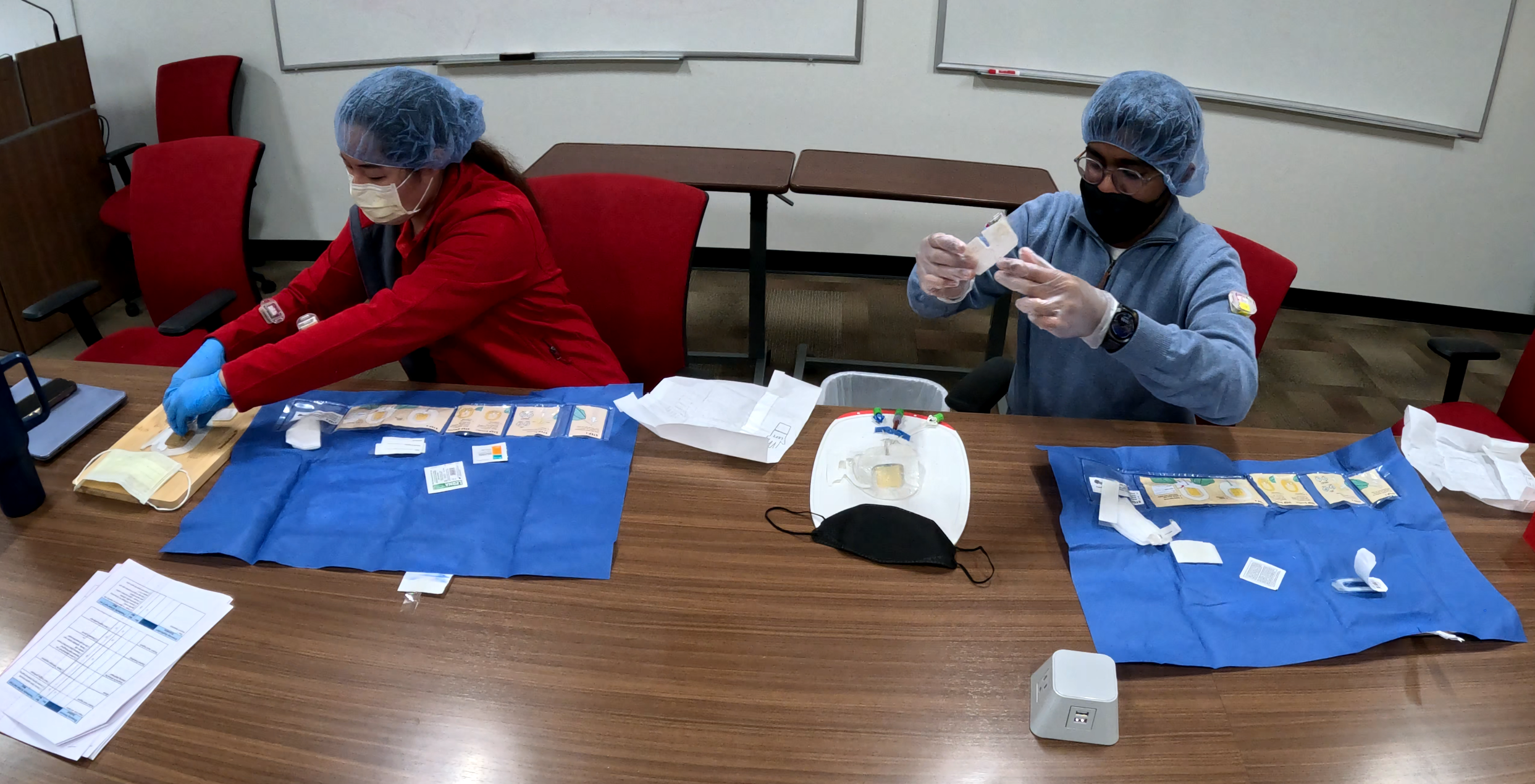}
    \caption{Training session for the data collection team prior to clinical deployment.}
    \label{fig:TrainingDataCollection}
\end{figure}

A formal training session was held on December 22, 2025, shown in Figure~\ref{fig:TrainingDataCollection}, with five team members (Gopi, Celestin, Spencer, Ummey, and Dr. Qian). 
During this session, Dr. Qian demonstrated the CLDC procedure while team members observed and practiced the recording workflow. 
The team rehearsed showing a pen to the camera at each activity transition to create visible markers for temporal segmentation. 
Dr. Pam and Ummey alternated performing the procedure to generate practice recordings from multiple operators, and the team practiced using two mannequin patients. 
This training ensured that all team members were familiar with both the clinical procedure and the recording protocol before data collection at Houston Methodist.

\subsection{Recording Workflow and On-Site Annotation}

\begin{figure}[H]
  \centering
  \includegraphics[width=0.49\columnwidth, height=3cm, keepaspectratio]{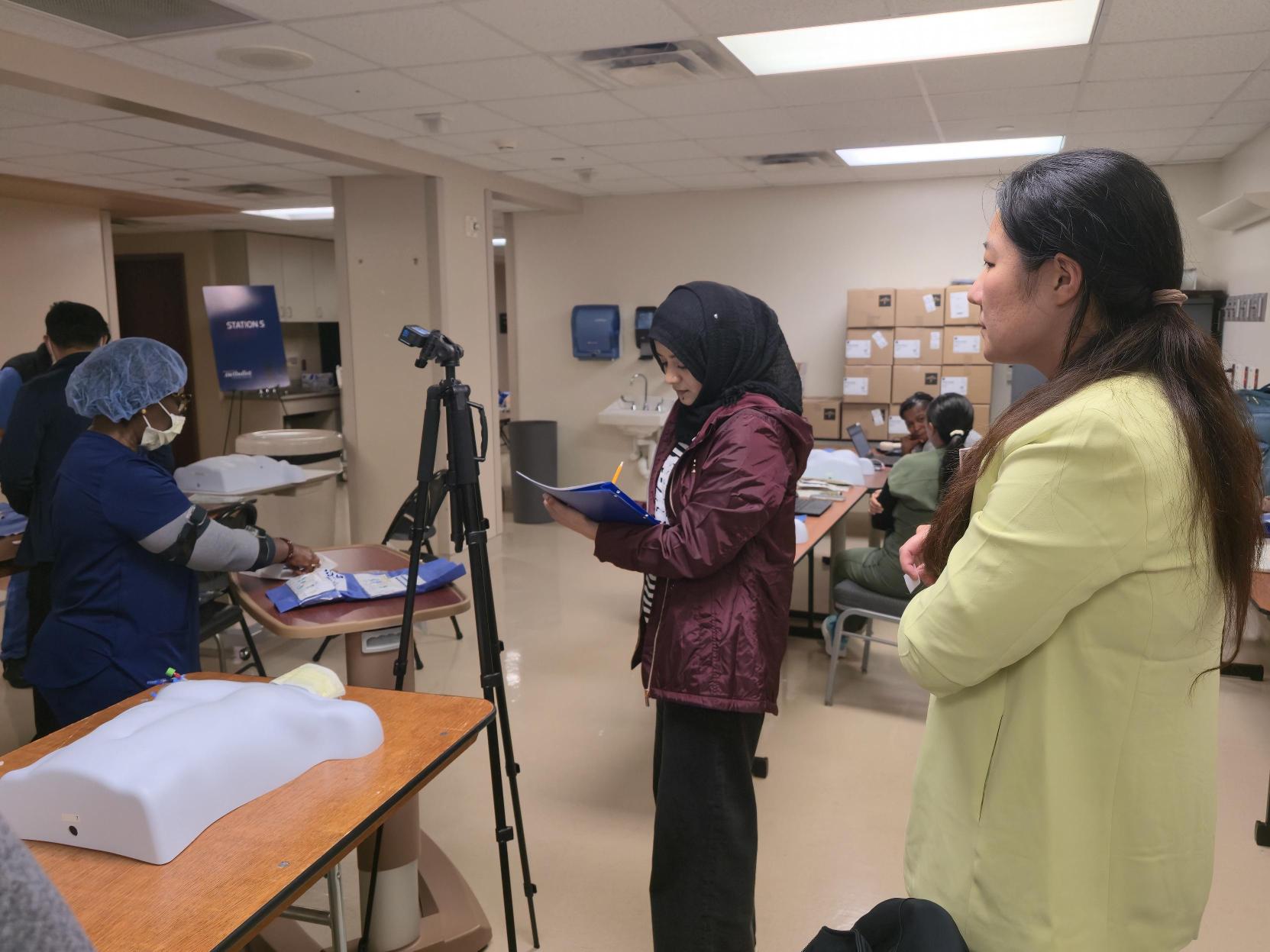}%
  \hspace{1pt}%
  \includegraphics[width=0.49\columnwidth, height=3cm, keepaspectratio]{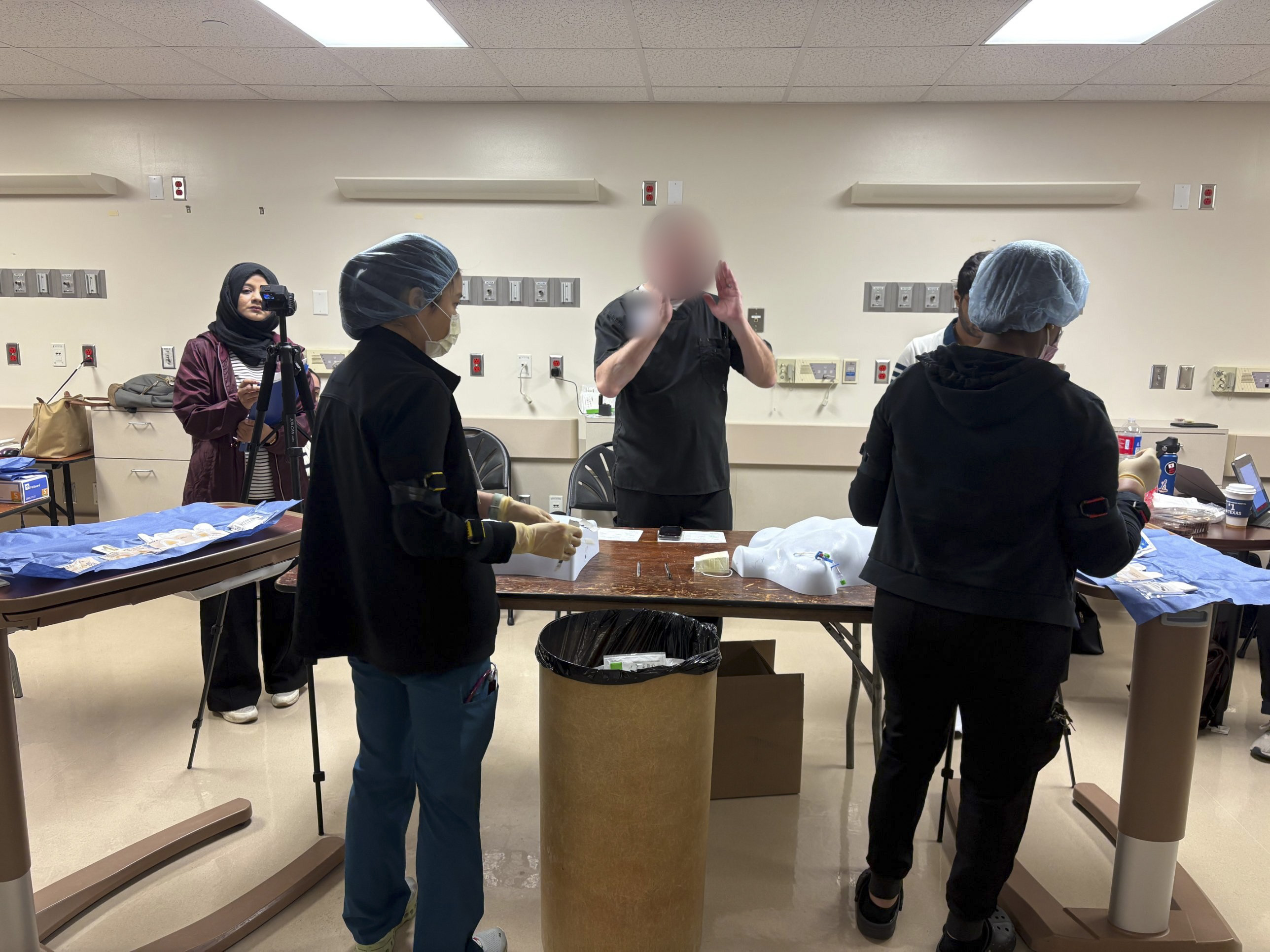}
  \caption{Real-time data annotation and nursing educator feedback during data collection. Featured nursing educator: Mr. Jay Hooker.}
  \label{fig:RealtimeAnnotation}
\end{figure}

Each participant was assigned a unique participant ID following a standardized format (e.g., \texttt{PID 001}). 
Printed ID numbers were shown to the camera at the start of each recording to enable unambiguous association between video files and participant records. 
Participants completed a demographic survey administered via QR code linked to a Microsoft Forms questionnaire. 
During each recording, a team of up to four trained observers performed concurrent data annotation, demonstrated in Figure~\ref{fig:RealtimeAnnotation}. 
Two observers focused on timekeeping by recording the start and end timestamps of each subtask activity using the finalized 17-activity checklist. 
The other two observers completed evaluation forms documenting qualitative feedback on the participant's performance. 
Printed hard copies of the evaluation forms were used rather than digital forms because the team determined that physical forms were more reliable and faster to complete during time-sensitive, real-time annotation. 
The demographic survey, being less time-critical, remained in digital format.

During real data collection, each recording session followed a standardized sequence: (1) explain the purpose of the study and the information to be collected, noting that only video of the hands and procedure area would be captured, with no face or audio recording intended; (2) assign the participant ID; (3) have the participant complete the demographic survey; (4) set up the camera and begin recording; (5) display the participant ID card to the camera; (6) allow the participant to perform the full CLDC procedure on the mannequin; (7) stop the camera upon procedure completion to conserve battery; and (8) complete the evaluation and timestamp forms.

\captionsetup{justification=centering}
\section{Data Records}
\label{sec:data-records}

\begin{figure*}[t]
  \centering
  \includegraphics[width=0.8\textwidth,keepaspectratio]{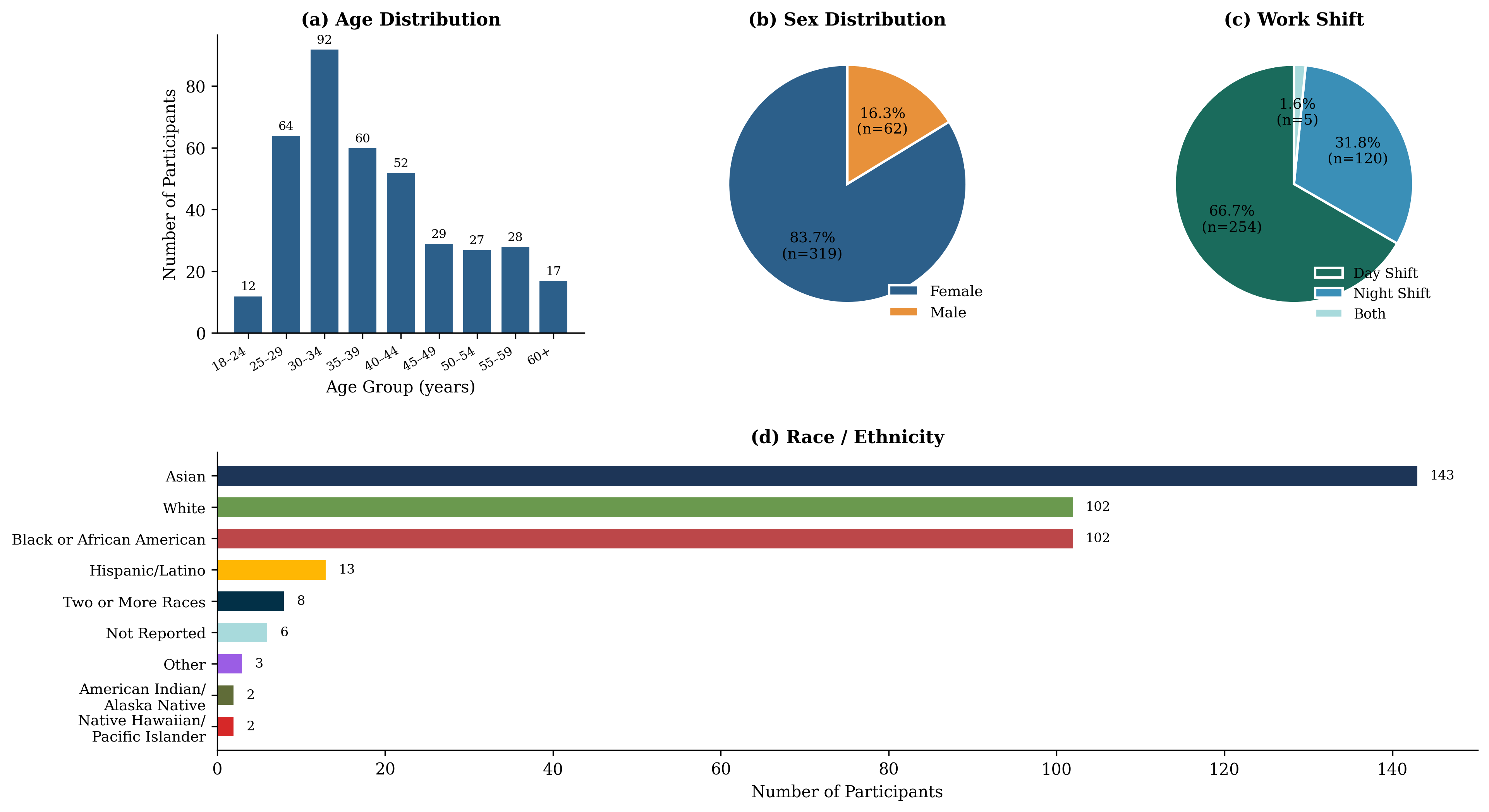}

  \includegraphics[width=0.8\textwidth,keepaspectratio]{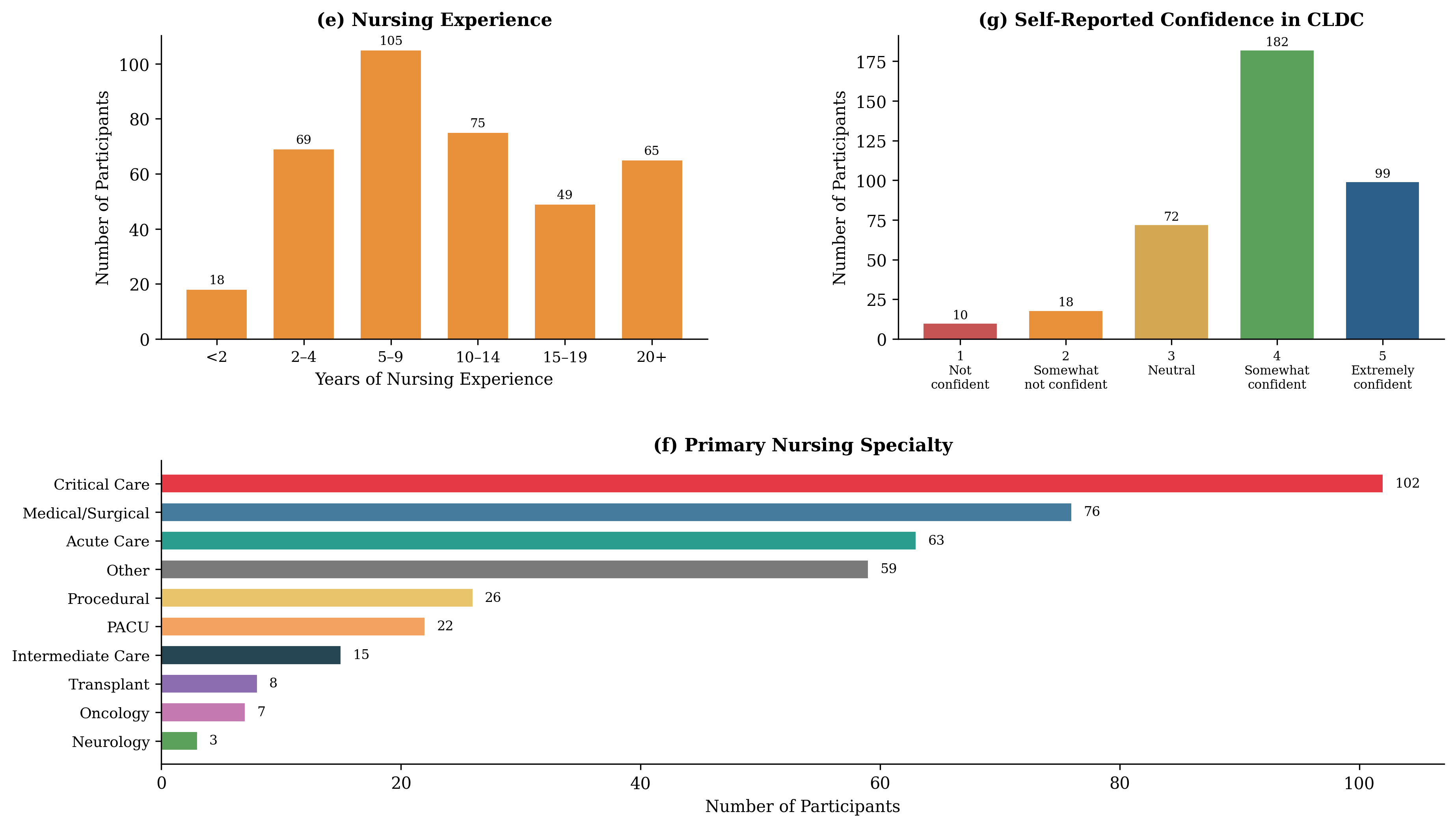}
  \caption{Participant demographics.}
  \label{fig:Demographics1}
\end{figure*}

The dataset consists of video recordings of the Central Line Dressing Change (CLDC) procedure collected at Houston Methodist Hospital during the hospital's annual January nursing competency evaluation, conducted over a three-week period from January 6 to January 22, 2026. 
During this evaluation, registered nurses from different departments rotated through designated competency stations where they performed clinical procedures on human torso mannequins. 
A total of 381 nurses participated in the study, each producing a single recorded video of the CLDC procedure. Of these, 16 recordings were excluded due to camera malfunctions, participant withdrawal mid-session, educator-initiated termination, or other unavoidable logistical interruptions. The remaining 365 complete recordings were retained for analysis (example video frames shown in Figure 9). 

As demonstrated in Figure~\ref{fig:Demographics1} (a)-(d), the majority of participants were female (83.7\%, $n=319$), with males comprising the remaining 16.3\% ($n=62$), reflecting the broader gender composition of the nursing workforce. 
Participants ranged in age from 22 to 70 years, with a mean age of 38.6 years (SD = 10.7), and the largest concentration in the 30--34 age group ($n=92$). 
The sample was racially and ethnically diverse: Asian nurses represented the largest group (37.5\%, $n=143$), followed by White (26.8\%, $n=102$) and Black or African American (26.8\%, $n=102$) participants, with smaller representation from Hispanic/Latino (3.4\%, $n=13$), two or more races (2.1\%, $n=8$), and other categories. 
Two-thirds of participants worked the day shift (66.7\%, $n=254$), while roughly one-third worked night shifts (31.8\%, $n=120$), and a small number reported working both (1.6\%, $n=5$).

In terms of professional background, Figure~\ref{fig:Demographics1} (e)-(g), participants brought a wide range of clinical experience to the study, with a mean of 11.4 years in nursing (SD = 9.1) and a median of 9 years. The largest group had 5--9 years of experience ($n=105$), although the dataset also includes nurses with fewer than 2 years of experience ($n=18$) as well as experienced professionals with over 20 years in the field ($n=65$). 
The most commonly reported nursing specialties were Critical Care (26.8\%, $n=102$), Medical/Surgical (19.9\%, $n=76$), and Acute Care (16.5\%, $n=63$), all of which are units where central line care is a routine component of practice. When asked to rate their confidence with the CLDC procedure, nearly half of the participants (47.8\%, $n=182$) described themselves as somewhat confident, and an additional 26.0\% ($n=99$) reported feeling extremely confident. 
A smaller proportion indicated neutral confidence (18.9\%, $n=72$), while only 7.3\% of participants reported low or no confidence, suggesting that the dataset primarily captures performance from nurses with at least some familiarity with the procedure rather than complete novices.

% \begin{figure*}[t]
%   \includegraphics[width=0.8\textwidth,keepaspectratio]{images/demographic_fig2.png}
%   \caption{Participant background characteristics: (a) nursing experience, (b) self-reported confidence in CLDC, and (c) primary nursing specialty.}
%   \label{fig:Demographics2}
% \end{figure*}

Recording equipment was deployed at two stations dedicated to the CLDC procedure. 
Each nurse performed the procedure individually in a standing position, consistent with standard bedside practice, and each recording captured the performance of a single registered nurse from start to finish. 
In total, we recorded approximately 381 videos across the three-week collection period, all at 4K resolution with a frame rate of 60 frames per second. Each video recording is paired with an associated annotation file following a consistent naming convention, for example, 
\texttt{GoProID\allowbreak\_\allowbreak PID009.mp4} for the video and \texttt{GoProID\_PID009.csv} for the annotation file. 
In this naming structure, \texttt{GoProID} denotes the GoPro internal file identifier and the number following \texttt{PID} denotes the unique participant identifier assigned at the time of recording. 
At the start of each video, a printed card displaying the participant ID was shown to the camera to provide a visual reference linking the recording to its corresponding participant and annotation records.

% \begin{figure}[H]
%   \centering
%   \includegraphics[width=7cm, height=4cm]{images/PID309_SpongeApplication1.PNG}

%   \vspace{4pt}

%   \includegraphics[width=7cm, height=4cm]{images/PID313-DRessingRemoval.png}
%   \caption{Example frames from the dataset. Top: sponge application by Participant 309. Bottom: dressing removal by Participant 313.}
%   \label{fig:ExampleFrames}
% \end{figure}

During each recording session, a member of the research team and a clinical nursing educator simultaneously completed a printed annotation form referred to as the CLDC Competency Validation Tool. 
The form is divided into two sections. The header captures session-level metadata, including the educator's name, validation date, station number, participant ID, and the overall recording start and end times. The body of the form contains one row for each subtask in the procedure. For every subtask, the annotator recorded the start time and end time relative to the beginning of the video. In parallel, the educator or annotation team member marked each subtask as either \textit{Met} or \textit{Not Met} according to whether the participant's performance satisfied the corresponding clinical criteria. Each row also includes predefined feedback descriptors, such as whether the clinician's mask was applied correctly, whether sterile technique was maintained, or whether the dressing was rolled rather than lifted toward the insertion site. These descriptors were checked or crossed out to indicate which quality indicators were or were not observed during performance.

\captionsetup{justification=centering}
\begin{figure}[H]
  \centering
  \includegraphics[width=6cm,keepaspectratio]{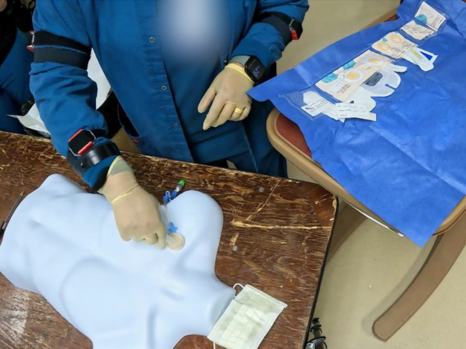}

  \vspace{4pt}

  \includegraphics[width=6cm, keepaspectratio]{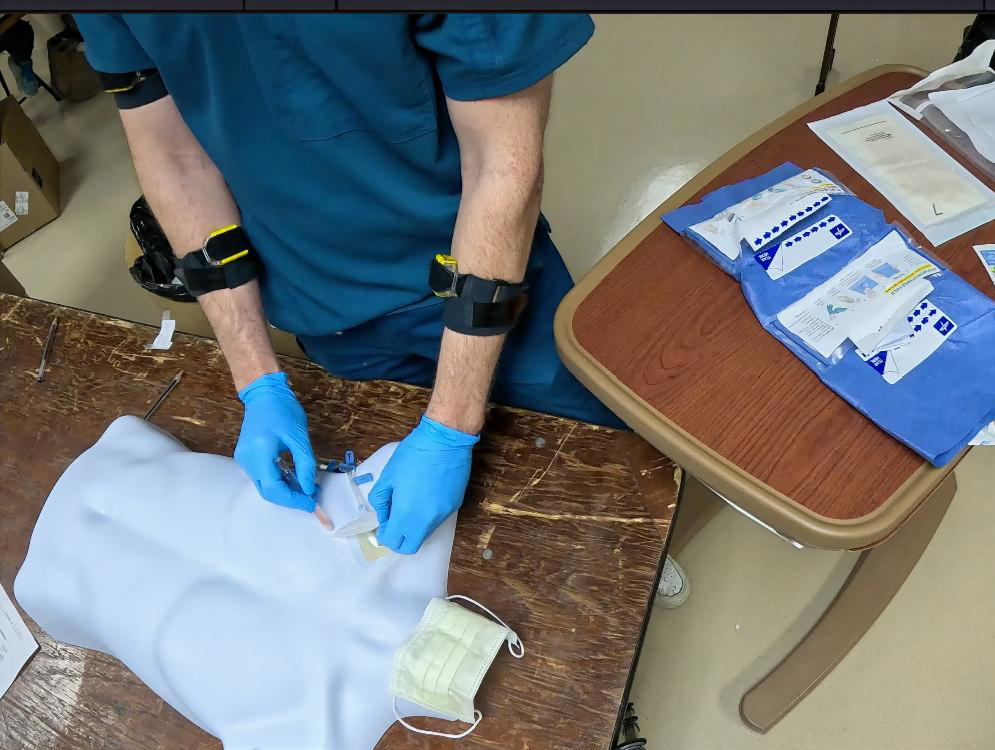}
  \caption{Example frames from the dataset. Top: sponge application by Participant 309. Bottom: dressing removal by Participant 313.}
  \label{fig:ExampleFrames}
\end{figure}

The printed annotation forms were subsequently digitized into CSV files. Each CSV file preserves the subtask-level start and end timestamps together with the \textit{Met}/\textit{Not Met} designations and any recorded feedback. These timestamps were verified against the video recordings to correct discrepancies introduced during live annotation. In cases where a nursing educator provided verbal guidance or hands-on instruction during a step, those teaching segments were documented so that they could be excluded from or flagged in downstream analysis.

\section{Conclusion and Future Work}
\label{sec:future-work}

This work makes three primary contributions. 
First, it reviews prior work related to CLDC-relevant data resources and skill assessment methods, spanning simulation-based nursing education and deep learning approaches for surgical skill evaluation, and identifies the absence of an automated system that directly addresses the central line dressing change procedure. 
Second, it presents a structured real-world data collection protocol for CLDC, including the systematic evaluation and selection of recording equipment, the development of annotation and evaluation instruments, and the successful collection of approximately 365 video recordings during the January 2026 nursing competency evaluation at Houston Methodist. 
To our knowledge, this dataset represents the first large-scale video collection of the CLDC procedure captured during authentic clinical competency assessment. 
Third, this work establishes an initial foundation for automated CLDC skill evaluation.

In the longer term, we are interested in exploring how AI-driven skill assessment can be integrated with robotic tutoring platforms such as QTrobot to support real-time coaching during practice sessions. In such a system, a nursing trainee could perform the CLDC procedure while the robot observes the procedure through a camera feed, recognizes subtasks as they unfold, evaluates the quality of execution, and delivers spoken feedback or demonstrations in real time. This direction could enable more accessible and consistent procedural practice by allowing trainees to receive immediate guidance even when one-to-one educator supervision is not available.

\begin{acks}
We thank the Center for Nursing Research, Education and Practice (CNREP) at Houston Methodist Hospital and its clinical educators for supporting on-site data collection, and the registered nurses who participated in this study. We thank Gopi Trinadh Maddikunta and Celestin Matala for their valuable assistance with the data collection process. 
\end{acks}

\balance
\bibliographystyle{ACM-Reference-Format}
\bibliography{main}

@article{Elangovan2024Impact,
    title={Impact of Central-Line-Associated Bloodstream Infections and Catheter-Related Bloodstream Infections: A Systematic Review and Meta-Analysis},
    author={Elangovan, S. and Lo, J. J. and Xie, Y. and Mitchell, B. and Graves, N. and Cai, Y.},
    journal={Journal of Hospital Infection},
    year={2024},
}

@article{Chovanec2021Association,
    title={Association of CLABSI With Hospital Length of Stay, Readmission Rates, and Mortality: A Retrospective Review},
    author={Chovanec, Kelli and Arsene, Camelia and Gomez, Cheryl and Brixey, Molly and Tolles, Diana and Wheaton Galliers, Jennifer and Kopaniasz, Renee and Bobash, Tiffany and Goodwin, Lamar},
    journal={Worldviews on Evidence-Based Nursing},
    year={2021},
}

@article{Star2024Reducing,
    title={Reducing Central Line-Associated Bloodstream Infection With a Dedicated CLABSI Prevention Registered Nurse Role},
    author={Star, Kelsey E. and Lindsey, Kaia and Reese, Sara M. and Paulino, Linda and Hernandez, Lily L. and Tynes, Irene and Eiseman, Karli and Tynes, John},
    journal={American Journal of Infection Control},
    year={2024},
}

@misc{Haddadin2023Central,
    title={Central Line Associated Blood Stream Infections},
    author={Haddadin, Yori and Annamaraju, Pavan and Regunath, Hariharan},
    year={2023},
    howpublished={\url{https://www.ncbi.nlm.nih.gov/books/NBK430891/}},
    note={StatPearls Publishing. Accessed: 2026-03-30},
}

@misc{AHRQ2024CLABSI,
    title={Central Line-Associated Bloodstream Infections (CLABSI)},
    author={{Agency for Healthcare Research and Quality}},
    year={2024},
    howpublished={\url{https://www.ahrq.gov/topics/central-line-associated-bloodstream-infections-clabsi.html}},
    note={Accessed: 2026-03-30},
}

@inproceedings{qian2025astrid,
  author={Qian, Peizhu and Bajraktari, Filip and Quintero-Pe\~{n}a, Carlos and Meng, Qingxi and Hamlin, Shannan and Kavraki, Lydia and Unhelkar, Vaibhav},
  title={{ASTRID}: A Robotic Tutor for Nurse Training to Reduce Healthcare-Associated Infections},
  booktitle={Proceedings of Robotics: Science and Systems (RSS)},
  year={2025},
  doi={10.15607/RSS.2025.XXI.090},
  url= {https://roboticsconference.org/program/papers/90/}
}

@inproceedings{huNurViD2023,
    author={Hu, Ming and Wang, Lin and Yan, Siyuan and Ma, Don and Ren, Qingli and Xia, Peng and Feng, Wei and Duan, Peibo and Ju, Lie and Ge, Zongyuan},
    title={{NurViD}: A Large Expert-Level Video Database for Nursing Procedure Activity Understanding},
    booktitle={Advances in Neural Information Processing Systems (NeurIPS 2023), Datasets and Benchmarks Track},
    volume={36},
    year={2023},
    url= {https://proceedings.neurips.cc/paper_files/paper/2023/hash/3a48b0eaba26ba862220a307a9edb0bb-Abstract-Datasets_and_Benchmarks.html},
}

@article{slamDataset2025,
    title={A Comprehensive Video Dataset for Surgical Laparoscopic Action Analysis},
    author={Ye, Zi and Zhou, Ru and Deng, Zili and Wang, Dan and Zhu, Ying and Jin, Xiaoli and Zhang, Lijun and Chen, Tianxiang and Zhang, Hanwei and Wang, Mingliang},
    journal={Scientific Data (Nature)},
    volume={12},
    pages={862},
    year={2025},
    doi={10.1038/s41597-025-05093-7},
}

@article{kcfResNet2023,
    title     = {An Automated Skill Assessment Framework Based on Visual Motion Signals and a Deep Neural Network in Robot-Assisted Minimally Invasive Surgery},
    author    = {Pan, Mingzhang and Wang, Shuo and Li, Jingao and Li, Jing and Yang, Xinjun and Liang, Kailin},
    journal   = {Sensors},
    volume    = {23},
    number    = {9},
    pages     = {4496},
    year      = {2023},
    doi       = {10.3390/s23094496},
}

@article{surgicalToolSurvey2022,
    title={Surgical Tool Datasets for Machine Learning Research: A Survey},
    author={Rodrigues, Mark and Mayo, Michael and Patros, Panos},
    journal={International Journal of Computer Vision},
    volume={130},
    number={9},
    pages={2222--2248},
    year={2022},
    doi={10.1007/s11263-022-01640-6},
}

@misc{surgVU2025,
    title={Surgical Visual Understanding ({SurgVU}) Dataset},
    author={Zia, Aneeq and Berniker, Max and Nespolo, Rogerio and Perreault, Conor and Wang, Ziheng and Mueller, Benjamin and Schmidt, Ryan and Bhattacharyya, Kiran and Liu, Xi and Jarc, Anthony},
    year={2025},
    eprint={2501.09209},
    archivePrefix={arXiv},
    primaryClass={cs.CV},
    url={https://arxiv.org/abs/2501.09209},
}

@article{DLSurgicalWorkflow2024,
    title={Deep Learning for Surgical Workflow Analysis: A Survey of Progresses, Limitations, and Trends},
    author={Li, Yunlong and Zhao, Zijian and Li, Renbo and Li, Feng},
    journal={Artificial Intelligence Review},
    volume={57},
    number={10},
    year={2024},
    doi={10.1007/s10462-024-10929-6},
}

@article{lavanchyAutomation2021,
    title={Automation of Surgical Skill Assessment Using a Three-Stage Machine Learning Algorithm},
    author={Lavanchy, Jo{\"e}l L. and Zindel, Jonas and Kirta{\c{c}}, Kadir and Twick, Ismail and Hosgor, Enes and Candinas, Daniel and Beldi, Guido},
    journal={Scientific Reports},
    volume={11},
    pages={5197},
    year={2021},
    doi={10.1038/s41598-021-84295-6},
}

@article{hashemiVideoRAS2025,
    title={Video-Based Robotic Surgical Action Recognition and Skills Assessment on Porcine Models Using Deep Learning},
    author={Hashemi, Nasseh and Mose, Matias and {\O}stergaard, Lasse R. and Bjerrum, Flemming and Hashemi, Maryam and Svendsen, Mikkel B. S. and Friis, Mikkel L. and Tolsgaard, Martin G. and Rasmussen, Sten},
    journal={Surgical Endoscopy},
    volume={39},
    number={3},
    pages={1709--1719},
    year={2025},
    doi={10.1007/s00464-024-11486-3},
}

@article{jamaNotss2024,
    title={Deep Learning Analysis of Surgical Video Recordings to Assess Nontechnical Skills},
    author={Harari, Rayan Ebnali and Dias, Roger D. and Kennedy-Metz, Lauren R. and Varni, Giovanna and Gombolay, Matthew and Yule, Steven and Salas, Eduardo and Zenati, Marco A.},
    journal={JAMA Network Open},
    year={2024},
    volume={7},
    number={7},
    pages={e2422520},
    doi={10.1001/jamanetworkopen.2024.22520},
}

@article{powerLSPD2025,
    title={Automated Assessment of Simulated Laparoscopic Surgical Skill Performance Using Deep Learning},
    author={Power, Daniel and Burke, Cathy and Madden, Michael G. and Ullah, Ihsan},
    journal={Scientific Reports},
    volume={15},
    pages={13591},
    year={2025},
    doi={10.1038/s41598-025-96336-5},
}

@article{sugiyamaMicro2025,
    title={Artificial Intelligence-Integrated Video Analysis of Vessel Area Changes and Instrument Motion for Microsurgical Skill Assessment},
    author={Sugiyama, Taku and others},
    journal={Scientific Reports},
    volume={15},
    year={2025},
    doi={10.1038/s41598-025-13522-1},
}

@article{frontiersSurgReview2025,
    title={Artificial Intelligence-Assisted Phase Recognition and Skill Assessment in Laparoscopic Surgery: A Systematic Review},
    author={Liao, Wenhao and Zhu, Ying and Zhang, Hanwei and Wang, Dan and Zhang, Lijun and Chen, Tianxiang and Zhou, Ru and Ye, Zi},
    journal={Frontiers in Surgery},
    volume={12},
    year={2025},
    pages={1551838},
    doi={10.3389/fsurg.2025.1551838},
}

@article{egocentricSkill2025,
    title={Egocentric Video Analysis for Automated Assessment of Open Surgical Skills via Deep Learning},
    author={Gazis, Alexandros and Schizas, Dimitrios and Kykalos, Stavros and others},
    journal={International Journal of Computer Assisted Radiology and Surgery},
    year={2025},
    doi={10.1007/s11548-025-03518-7},
}

@article{bridgingGapBMC2025,
    title     = {The Impact of Clinical Simulation on Bridging the Theory--Practice Gap in Nursing Education: A Systematic Review},
    author    = {Daneshfar, Majid and Karimi Moonaghi, Hossein},
    journal   = {BMC Medical Education},
    volume    = {25},
    pages     = {1216},
    year      = {2025},
    doi       = {10.1186/s12909-025-07790-8},
}

@article{sblSystematic2024,
    title     = {The Effectiveness of Simulation-Based Learning ({SBL}) on Students' Knowledge and Skills in Nursing Programs: A Systematic Review},
    author    = {Alharbi, Ali and Nurfianti, Arina and Mullen, Rosemary F. and McClure, John D. and Miller, William H.},
    journal   = {BMC Medical Education},
    volume    = {24},
    pages     = {1099},
    year      = {2024},
    doi       = {10.1186/s12909-024-06080-z},
}

@article{chabreraOSCE2023,
    title     = {Development, Validity and Reliability of Objective Structured Clinical Examination in Nursing Students},
    author    = {Chabrera, Carolina and Diago, Eva and Curell, Laura},
    journal   = {SAGE Open Nursing},
    volume    = {9},
    year      = {2023},
    doi       = {10.1177/23779608231207217},
}

@article{osceKorea2023,
    title     = {The Effect of Objective Structured Clinical Examinations for Nursing Students},
    author    = {Ha, Eun-Ho and Lim, Eunhee},
    journal   = {PLOS ONE},
    volume    = {18},
    number    = {6},
    pages     = {e0286787},
    year      = {2023},
    doi       = {10.1371/journal.pone.0286787},
}

@article{simDesignCSN2024,
    title     = {Impact of Simulation Design Elements on Undergraduate Nursing Education: A Systematic Review},
    author    = {Jackson, Meredith and McTier, Lauren and Brooks, Lucy A. and Wynne, Robyn},
    journal   = {Clinical Simulation in Nursing},
    volume    = {89},
    pages     = {101519},
    year      = {2024},
    doi       = {10.1016/j.ecns.2024.101519},
}

@article{vrScopingCSN2024,
    title     = {Exploring the Use of Immersive Virtual Reality in Nursing Education: A Scoping Review},
    author    = {Bradley, Cassandra S. and Wilson, Jennifer and Thompson, Desiree},
    journal   = {Clinical Simulation in Nursing},
    volume    = {97},
    pages     = {101648},
    year      = {2024},
    url       = {https://www.nursingsimulation.org/article/S1876-1399(24)00140-3/fulltext},
}

@article{osceStressPLOS2025,
    title     = {Exploring the Challenges and Needs of Nursing Students in Relation to {OSCE} Exam Stress: A Qualitative Study},
    author    = {Hosseini, Fahimeh Alsadat and Hemati, Mehrvash and Shaygan, Maryam and Gheysari, Somayeh and Jaberi, Azita and Ghobadi, Mojtaba and Rahimi, Shoeleh},
    journal   = {PLOS ONE},
    volume    = {20},
    number    = {7},
    pages     = {e0327898},
    year      = {2025},
    doi       = {10.1371/journal.pone.0327898},
}

@article{sibleyOSCE2025,
    title     = {Using Unfolding {OSCEs} Across a Graduate Nursing Curriculum to Implement Competency-Based Education Assessment},
    author    = {Sibley, Sean and Sajedi Sabegh, Nima and Herbert, Valerie},
    journal   = {Clinical Simulation in Nursing},
    year      = {2025},
    url       = {https://www.sciencedirect.com/science/article/abs/pii/S187613992500088X},
}

@inproceedings{funke2019video,
  title={Video-based surgical skill assessment using 3D convolutional neural networks},
  author={Funke, Isabel and Mees, S{\"o}ren Torge and Weitz, J{\"u}rgen and Speidel, Stefanie},
  booktitle={International Journal of Computer Assisted Radiology and Surgery},
  volume={14},
  number={7},
  pages={1217--1225},
  year={2019},
  publisher={Springer}
}

@article{gorski2021,
  author    = {Gorski, Lisa A. and Hadaway, Lynn and Hagle, Mary E. and Broadhurst, Dawn and Clare, Sharon and Kleidon, Tricia and Meyer, Britt M. and Nickel, Barb and Rowley, Simon and Sharpe, Elaine and Alexander, Mary},
  title     = {Infusion Therapy Standards of Practice},
  journal   = {Journal of Infusion Nursing},
  volume    = {44},
  number    = {1S Suppl 1},
  pages     = {S1--S224},
  year      = {2021},
  edition   = {8th},
  doi       = {10.1097/NAN.0000000000000396}
}
\balance

\end{document}